\documentclass{article}
\usepackage{ijcai26}

\usepackage{times}
\usepackage{soul}
\usepackage{url}
\usepackage[hidelinks]{hyperref}
\usepackage[utf8]{inputenc}
\usepackage[small]{caption}
\usepackage{graphicx}
\usepackage{svg}
\usepackage{amsmath}
\usepackage{amssymb}
\usepackage{amsthm}
\usepackage{tabularx}
\usepackage{booktabs}
\usepackage{algorithm}
\usepackage{algorithmic}
\usepackage[switch]{lineno}
\usepackage{enumitem}
\usepackage{pgfplots}
\usepackage{cleveref}
\usepackage{wrapfig}
\usepackage{svg}
\usepackage{pgfplots}
\usepackage{pgfplotstable}
\usepackage{subcaption}
\usetikzlibrary{patterns}
\newcolumntype{R}{>{\raggedleft\arraybackslash}X}

\usepackage{lipsum}

\usepackage{xcolor}
\usepackage{chato-notes} 

\newcommand{\citet}[1]{{\citeauthor{#1}~\shortcite{#1}}}

\pgfplotsset{compat=1.17}

\definecolor{col5}{HTML}{009933} 
\definecolor{col4}{HTML}{99ff99} 
\definecolor{col3}{HTML}{ffff00} 
\definecolor{col2}{HTML}{ff9900} 
\definecolor{col1}{HTML}{ff0000} 

\pgfplotsset{
    ruva_barplot/.style={
        xbar stacked,
        width=\textwidth,
        height=4.5cm,
        bar width=12pt,
        xmin=0, xmax=100,
        
        table/col sep=comma,
        
        ytick=data,
        yticklabels from table={#1}{Model},
        y dir=reverse,
        xticklabel={\pgfmathprintnumber\tick\%},
        xlabel={}, 
        
        y tick label style={font=\scriptsize\sffamily},
        x tick label style={font=\tiny\sffamily},
        
        legend style={
            at={(0.5,-0.4)},
            anchor=north,
            legend columns=-1,
            draw=none,
            font=\tiny\sffamily,
            /tikz/every even column/.append style={column sep=0.2cm}
        },
        nodes near coords={%
            \pgfkeys{/pgf/fpu=true}%
            \pgfmathparse{\pgfplotspointmeta<4 ? "" : \pgfmathprintnumber{\pgfplotspointmeta}\%}%
            \pgfmathresult%
        },
        every node near coord/.append style={
            font=\tiny\sffamily,
            color=black,
            yshift=0pt
        },
        axis line style={draw=none},
        tick style={draw=none},
        xmajorgrids=true,
        grid style={dotted, gray!30}
    }
}

\title{RUVA: Personalized Transparent On-Device Graph Reasoning}

\author{
Gabriele Conte$^1$ \and
Alessio Mattiace$^1$\and
Gianni Carmosino$^1$\and\\
Potito Aghilar$^1$\and
Giovanni Servedio$^1$\and
Francesco Musicco$^1$\and\\
Vito Walter Anelli$^1$\and
Tommaso Di Noia$^1$\And
Francesco Maria Donini$^2$
\affiliations
$^1$Politecnico di Bari, 
$^2$Università degli Studi della Tuscia\\
\emails
\{g.conte12, a.mattiace, g.carmosino1\}@studenti.poliba.it,\\ \{potito.aghilar, giovanni.servedio, vitowalter.anelli, tommaso.dinoia\}@poliba.it\\
f.musicco@phd.poliba.it, donini@unitus.it
}

\begin{document}

\maketitle

\begin{abstract}
The Personal AI landscape is currently dominated by ``Black Box" Retrieval-Augmented Generation. While standard vector databases offer statistical matching, they suffer from a fundamental lack of accountability: when an AI hallucinates or retrieves sensitive data, the user cannot inspect the cause nor correct the error. Worse, ``deleting" a concept from a vector space is mathematically imprecise, leaving behind probabilistic ``ghosts" that violate true privacy. We propose \textbf{\textsc{Ruva}}, the first ``Glass Box" architecture designed for Human-in-the-Loop Memory Curation. 
Ruva grounds Personal AI in a Personal Knowledge Graph, enabling users to inspect \textit{what} the AI knows and to perform precise redaction of specific facts. By shifting the paradigm from Vector Matching to Graph Reasoning, \textsc{Ruva} ensures the ``Right to be Forgotten." Users are the editors of their own lives; \textsc{Ruva} 
hands them the pen.



\end{abstract}

\section{Introduction}
Since their release, Large Language Models have quickly transitioned from novelties to commodities~\cite{DBLP:journals/patterns/LiangZCWCZ25}. The early industry dream of hyper-personalizing these models via fine-tuning has collapsed under the weight of computational costs and the risk of catastrophic forgetting~\cite{luo2025empirical}. Consequently, the field has converged on Retrieval-Augmented Generation (RAG)~\cite{DBLP:conf/nips/LewisPPPKGKLYR020} as the only viable path for personalization. However, the reliance on standard Vector RAG has created a crisis: the ``Black Box" problem.
Current RAG systems compress a user's life (e.g., emails, photos, chats, calendar entries) into high-dimensional embedding spaces. While efficient for matching semantically similar keywords, the user cannot inspect why the system fails or fix it.
More critically, true privacy requires the ability to delete, but one cannot surgically remove a concept from a vector space; one can only suppress it. This mathematical imprecision leaves behind ``probabilistic ghosts", traces of sensitive data that the AI might accidentally resurrect~\cite{Wang2025}.
GraphRAG methods~\cite{Edge} address these limitations by incorporating graph topology into retrieval.

We introduce \textsc{Ruva} (the Etruscan word for ``Brother"), a system designed to be as trustworthy and protective as a close sibling. \textsc{Ruva} is a \textbf{neuro-symbolic GraphRAG architecture}~\cite{Kautz22} that runs entirely \textbf{on-device}, representing a paradigm shift from opaque retrieval to transparent reasoning.
Trust begins with sovereignty. By running locally, \textsc{Ruva} ensures that the user's Personal Knowledge Graph (PKG)~\cite{BalogK19,SkjaevelandBBLL24} never leaves their device. There is no cloud to hack; the user's life remains physically in their pocket. Unlike vector stores, \textsc{Ruva} grounds its intelligence in a ``Glass Box" architecture. Here, every memory is an explicit node, and every association is a visible edge. This structure highlights a core consideration: \textit{Vectors allow Matching; Graphs allow Reasoning.}
Consider the query: \textit{``Did Sarah call before I arrived at work?"} A standard vector system simply retrieves chunks containing ``Sarah" and ``Work" based on similarity, often hallucinating the temporal relationship. In contrast, \textsc{Ruva} traverses the graph topology, comparing the timestamp of the \textit{Call} node against the \textit{Arrival} node to provide the answer.

Finally, \textsc{Ruva} redefines the user's role from a passive data source to the \textit{Editor-in-Chief}. Because the memory is structured, it is fully editable. If \textsc{Ruva} learns something incorrect or sensitive, the user can use the \textit{``Red Pen"} to perform a precise redaction, cutting that specific node (the reified information object) and its edges. The knowledge is deleted instantly and mathematically, ensuring the ``Right to be Forgotten."
\textsc{Ruva} is not another chatbot. It is an AI that users can actually trust, because \textsc{Ruva} allows them to see inside its head and edit its memories. The project and the demo video are available at \url{http://sisinf00.poliba.it/ruva/}.

\begin{figure*}[t]
    \centering
    \includegraphics[height=15em]{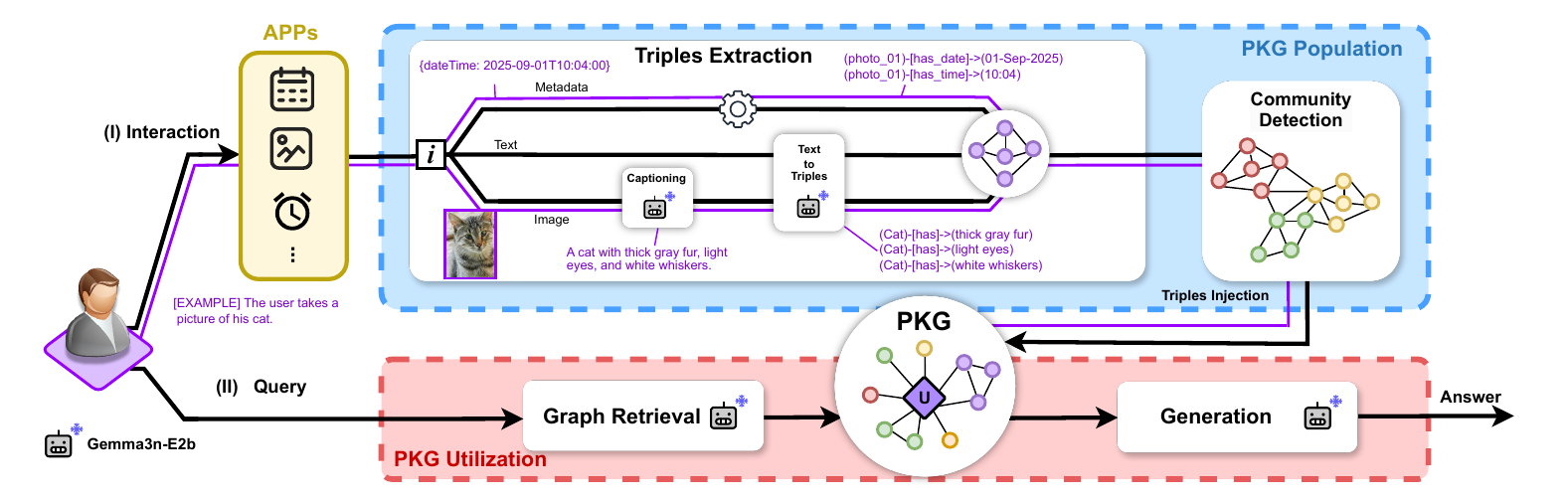}
    \caption{The \textsc{Ruva} architecture. The \textit{Ingestion Workflow} transforms multimodal data into semantic triples to populate the Personal Knowledge Graph. The \textit{Retrieval Workflow} performs graph traversal to generate grounded, hallucination-free answers, entirely on-device.}
    \label{fig:architecture}\vspace{-1em}
\end{figure*}

\section{The \textsc{Ruva} Architecture}
\label{sec:architecture}

\textsc{Ruva} is engineered as an offline-first, mobile-native system designed to guarantee data sovereignty. To ensure that personal data never leaves the user's possession, all computation occurs locally on the user's mobile device; the \textbf{Personal Knowledge Graph (PKG)} 
is physically bound to the handset, eliminating cloud-based attack vectors. Furthermore, the system is optimized for mobile constraints, utilizing quantized models and efficient storage to operate alongside the OS without draining the battery.
As illustrated in Figure~\ref{fig:architecture}, \textsc{Ruva} implements a \textbf{Type~3 Neuro-Symbolic architecture} (taxonomy from~\citet{Kautz22}) that separates perception from memory. The \textit{Neural Component} employs a Small Vision Language Model (SVLM, i.e., Gemma-3n-E2b~\cite{Gemma3}) as the system's ``eyes and ears." Its role is to convert unstructured noise (pixels, raw text) into structured signals, accepting the probabilistic nature of recognition. These signals are fed into the \textit{Symbolic Component}, a PKG used for storage and reasoning (\Cref{fig:app_pkg}). Unlike the neural component, the memory, once created, is deterministic, inspectable, and editable.
The system backbone is a hybrid storage engine: a single SQLite database extended with \texttt{sqlite-vec}. For this research, the engine is extended to accommodate GraphRAG~\cite{Edge} operations. This design choice is critical for the ``Glass Box" paradigm.
The resulting graph topology resembles a ``spiderweb," where a central \texttt{User} node acts as the root, and all entities, \texttt{Events}, \texttt{Photos}, \texttt{Messages}, branch out through typed edges, creating a unified representation of the user's digital footprint.

\subsection{Multimodal Ingestion Workflow}
\textsc{Ruva} operates a background ``Triple Extraction" pipeline that transforms heterogeneous data streams into semantic triples. For visual data, such as a photo of a train ticket, the image is passed to the local SVLM. The SVLM generates a dense caption, which is then parsed to extract structured entities, effectively converting pixels into nodes like \texttt{(:Receipt)-[:cost]->("95 EUR")}. Textual data, such as emails or notes, is processed directly by the SVLM to extract entities and temporal metadata. To prevent graph fragmentation, \textsc{Ruva} employs an Entity Resolution strategy along with community detection (Leiden algorithm~\cite{Traag_2019}). For instance, if the system ingests an email from ``Sarah Green" and a calendar invite from ``S. Green," the clustering identifies these as the same entity, merging them into a single \texttt{Person} node rather than creating duplicates. 

\subsection{Graph-Grounded Retrieval Workflow}
\begin{figure}[h!]
    \centering
    
    \begin{subfigure}[t]{0.49\columnwidth}
        \centering
        \includegraphics[height=21em]{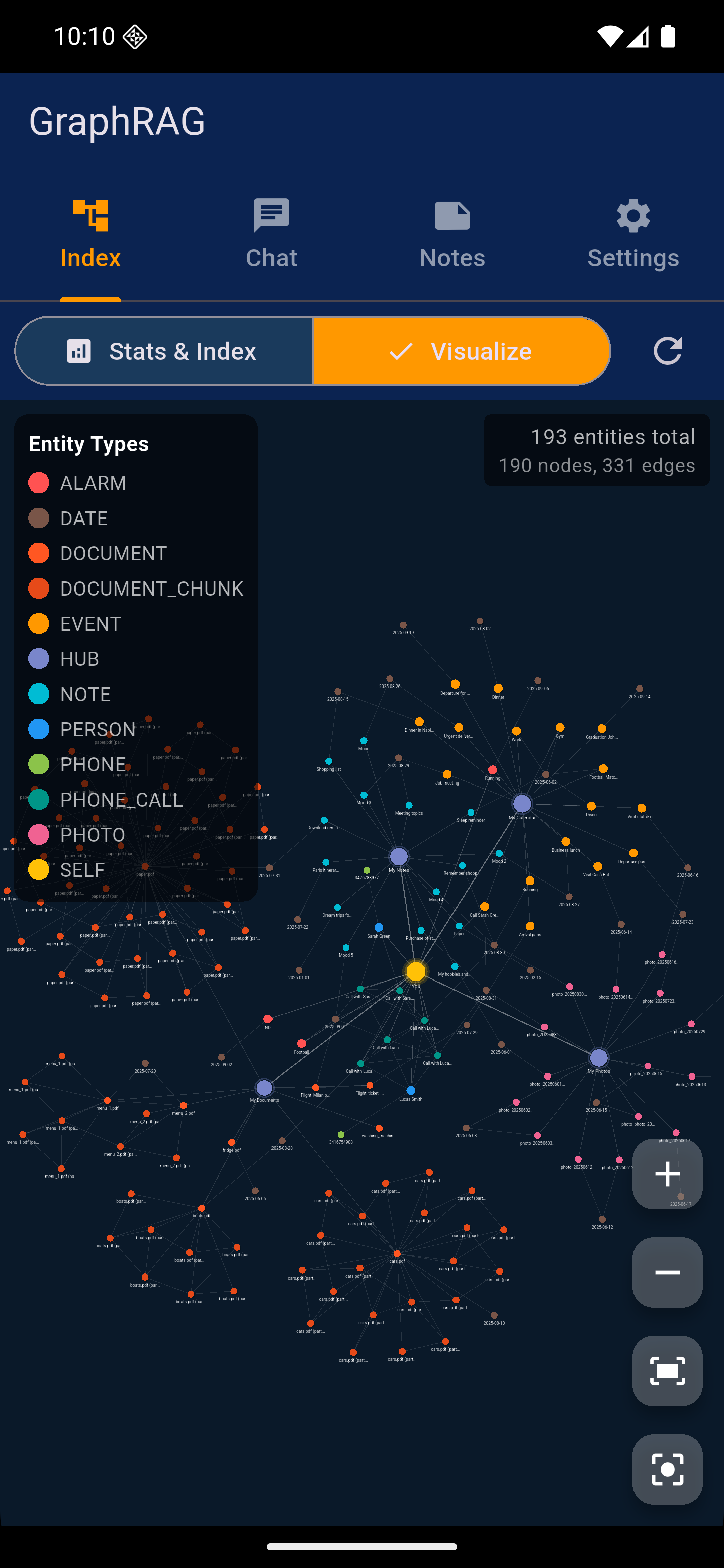}
        \caption{Extracted PKG in \textsc{Ruva}}\label{fig:app_pkg}
    \end{subfigure}
    \hfill
    \begin{subfigure}[t]{0.49\columnwidth}
        \centering
        \includegraphics[height=21em]{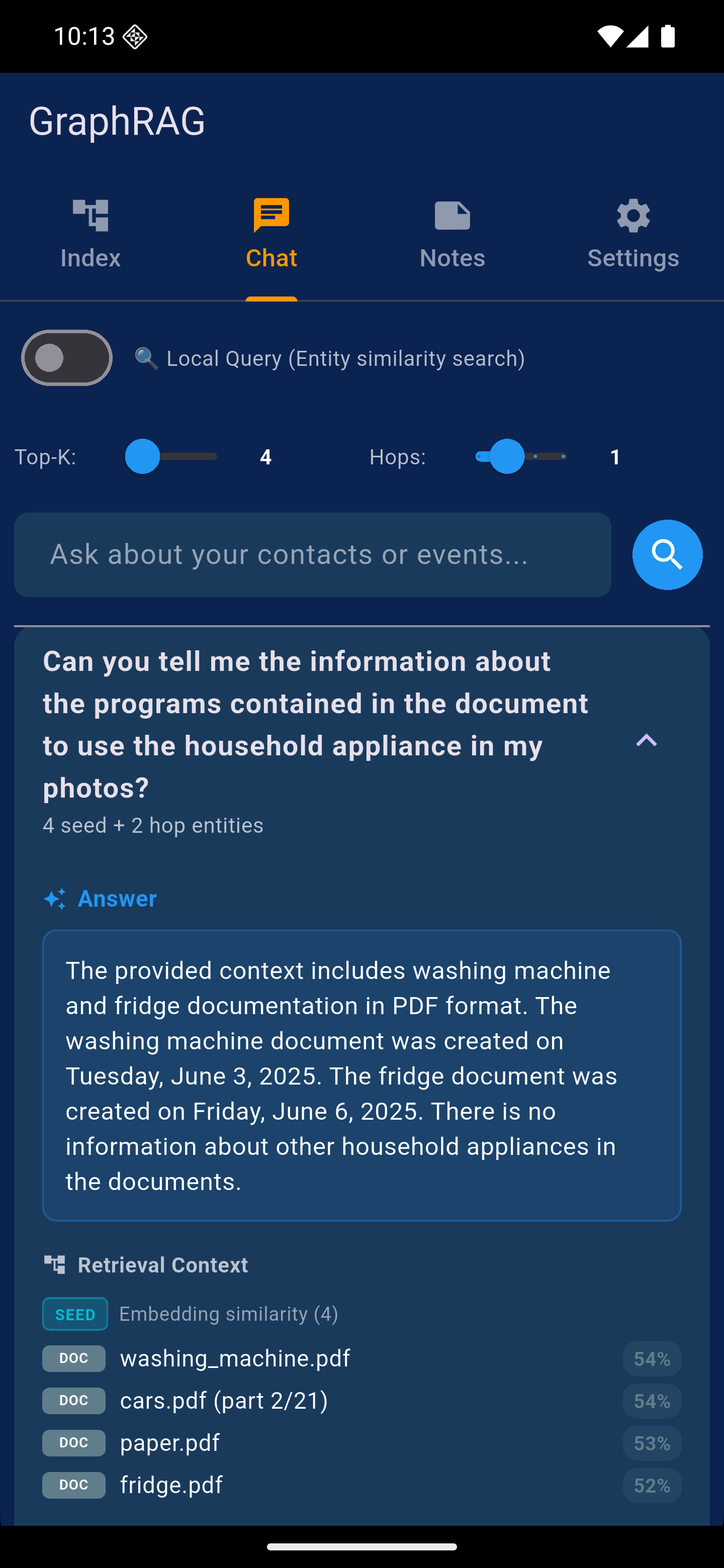}
        \caption{Answer to multimodal query}\label{fig:app_complex_query}
    \end{subfigure}
    
    \caption{\textsc{Ruva} App Screenshots. \Cref{fig:app_pkg} shows the inferred PKG, while in  \Cref{fig:app_complex_query} \textsc{Ruva} answers leveraging multimodal data.}\vspace{-1em}
\end{figure}

Standard Vector RAG fails at complex reasoning because it relies solely on semantic similarity. \textsc{Ruva} solves this via a \textbf{multi-step Graph-Grounded Retrieval mechanism}. Upon receiving a query like \textit{``Did Sarah call before I arrived at work?"}, the system first executes an \textbf{Anchor Search} using vector similarity to find the relevant ``Anchor Nodes" (e.g., the \texttt{Person:Sarah} node and the \texttt{Location:Work} node). It then performs a configurable \textbf{Topological Expansion}, executing an $N$-hop traversal from these anchors to identify connected events. Finally, during \textbf{Answer Generation} (\Cref{fig:app_complex_query}), the retrieved subgraph, containing explicit timestamps and relationships, is serialized and fed into the SVLM. 

\subsection{The Deletion Mechanism}
One of the most significant advantages of the \textsc{Ruva} architecture is user-controlled knowledge deletion. In vector-only systems, deleting a concept is mathematically imprecise, leaving behind probabilistic ghosts. In \textsc{Ruva}, since every memory is a discrete node in a relational database, deletion is exact. When a user chooses to forget a specific memory (e.g., ``The project meeting on Friday"), \textsc{Ruva} executes a standard SQL \texttt{DELETE CASCADE} transaction targeting the reified information object. The node, its properties, and all associated vector indices are instantly and permanently excised from the memory. This provides the user with deterministic control over their digital past.

\section{Demonstration Scenarios}
We demonstrate the ``Glass Box" \textsc{Ruva} architecture with three use cases: (i) ingestion of a multimodal memory; (ii) complex reasoning, and (iii) memory deletion.

The demonstration begins with \textbf{Multimodal Ingestion (Scenario 1)}, illustrating how the system reacts when multimodal data enters the system.
Let us introduce an example. The user syncs a calendar containing a ``Weekend Trip'' and uploads a photo of a train ticket. The \textit{Multimodal Ingestion Workflow} triggers: the local VLM converts the ticket pixels into a structured node (\texttt{:Receipt :amount "95~EUR"}), while the system identifies the temporal overlap with the calendar event, forging a relationship edge.

Next, we demonstrate the \textbf{Complex Reasoning (Scenario~2)}. Again, let us show it through an example. The user asks: \textit{``How much have I spent on the trip so far?"} Answering this requires bridging the concept ``The Trip" from the calendar with granular data ``95 EUR" from the image. The GraphRAG runtime executes a multi-hop traversal, locating the Trip node and following the temporal edge to the Receipt node, answering \textit{``You have spent 95 EUR for the ticket."}

Finally, we showcase the \textbf{Deterministic Deletion (Scenario~3)}, addressing the ``Right to be Forgotten.'' In this example, the user decides that a financial piece of information (i.e., the receipt in Scenario 2) is sensitive and should not be retained. By selecting the \textit{Receipt} in the gallery app and removing it, the system executes a deterministic \texttt{DELETE} cascade transaction. The node and all associated vector indices are excised. To verify this, the user asks the same question again, and the system responds:
\textit{``I couldn't find relevant information to answer your question."} This scenario demonstrates true data sovereignty: the user acts as the editor of their own memory, enforcing privacy with mathematical certainty in a way that opaque vector-only systems cannot guarantee.

\section{Evaluation}
\label{sec:evaluation}
 
This section assesses the \textsc{Ruva}'s \textit{operational feasibility} of the architecture and the \textit{semantic reasoning quality}.
 
\noindent \textbf{Performance.}
We deployed the system on a constrained environment, a Google Pixel 8 Pro. \textsc{Ruva} periodically scans the device's local data and compares it with existing nodes. 
Despite the constrained environment, \textsc{Ruva} ensures interactive latencies, with an average ingestion time of \textit{2.4s} (including VLM processing and triple extraction) and a graph retrieval latency of only \textit{38ms} for single-hop queries. 
 
\noindent \textbf{Model Accuracy.}
To comprehensively test \textsc{Ruva}, we prepared a benchmark comprising \textit{71} multi-source objects (spanning calendar events, images, notes, textual documents, calls, alarms, and contacts) and $52$ triplets containing questions, desired answers, and \textsc{Ruva}'s answers. The benchmark specifically probes three distinct scenarios: \textit{Multimodal Ingestion} (\Cref{fig:image1}), \textit{Complex Reasoning} (\Cref{fig:image2}), and \textit{Deterministic Deletion} (\Cref{fig:image3}). 
The benchmark consists of \textbf{20 ingestion checks} and \textbf{32 complex reasoning} questions.
To assess \textit{Deterministic Deletion}, each question is evaluated with ($score_0$) and without ($score_1$) its corresponding information objects, measuring $\Delta = score_0 - score_1$. We considered answers whose judges' score is positive (3-5).
 
{\noindent \textbf{Experiments.}
We employ a panel of four state-of-the-art LLMs: Llama3.3-70B~\cite{grattafiori2024llama3herdmodels}, Qwen3-32B~\cite{yang2025qwen3technicalreport}, GPT~\cite{openai2025gptoss}, and Kimi K2~\cite{kimiteam2025kimik2openagentic} Instruct.
Following the Prometheus~\cite{prometheusinducingfinegrainedevaluation} strategy, judges evaluate benchmark triplets via Chain-of-Thought reasoning.
To mitigate inter-model bias~\cite{lee-etal-2025-checkeval}, we map the resulting 5-point scores to three categories: \textit{Positive} (4–5), \textit{Neutral} (3), and \textit{Negative} (1–2).
\textsc{Ruva} demonstrates robust reasoning, with \textbf{61\%} of responses rated as \textbf{Positive}, increasing to \textbf{71\%} when including \textbf{Neutral} responses. (\Cref{fig:image1}, \Cref{fig:image2}). Reliability analysis shows strong cross-model agreement, with a Spearman’s Rank Correlation ($\rho$)~\cite{Spearman1904Proof} of \textbf{0.82} and a Krippendorff's Alpha~\cite{Krippendorff2011ComputingKA} of \textbf{0.81}, indicating consistency in ordinal evaluations. The \textbf{83\% Percentage Agreement} ~\cite{Interraterreliability} and \textbf{Cohen’s $\kappa$}~\cite{Cohen1968Weighted} of \textbf{0.81} confirm this finding. Notably, \texttt{gpt-oss-120b} and \texttt{kimi-k2-instruct-0905} achieve up to \textbf{92\% agreement} ($\kappa=0.92$), highlighting highly aligned assessments.

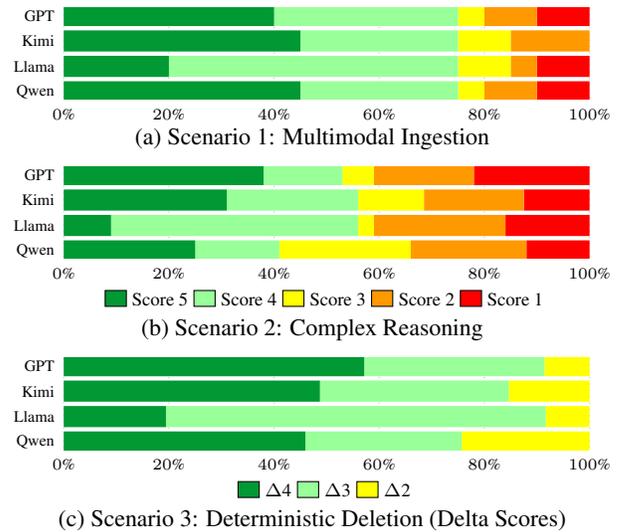
\begin{figure}
    \centering

    \begin{subfigure}{\columnwidth}
        \centering
        \begin{tikzpicture}
            \begin{axis}[
                xbar stacked,
                width=\textwidth, height=2.8cm,
                bar width=8pt,
                xmin=0, xmax=100,
                ytick=data,
                yticklabels from table={plots/Scen1.csv}{Model},
                table/col sep=semicolon,
                y dir=reverse,
                xticklabel={\pgfmathprintnumber{\tick}\%},
                xticklabel style={font=\tiny},
                yticklabel style={font=\tiny},
                axis line style={draw=none},
                tick style={draw=none},
                xmajorgrids=true,
                grid style={dotted, gray!50},
            ]
                \addplot[fill=col5, draw=none] table [x=S5, y expr=\coordindex] {plots/Scen1.csv};
                \addplot[fill=col4, draw=none] table [x=S4, y expr=\coordindex] {plots/Scen1.csv};
                \addplot[fill=col3, draw=none] table [x=S3, y expr=\coordindex] {plots/Scen1.csv};
                \addplot[fill=col2, draw=none] table [x=S2, y expr=\coordindex] {plots/Scen1.csv};
                \addplot[fill=col1, draw=none] table [x=S1, y expr=\coordindex] {plots/Scen1.csv};
            \end{axis}
        \end{tikzpicture}\vspace{-0.7em}
        \caption{Scenario 1: Multimodal Ingestion}\label{fig:image1}
    \end{subfigure}\vspace{0.2em}

    \begin{subfigure}{\columnwidth}
        \centering
        \begin{tikzpicture}
            \begin{axis}[
                xbar stacked,
                width=\textwidth, height=2.8cm,
                bar width=8pt,
                xmin=0, xmax=100,
                ytick=data,
                yticklabels from table={plots/Scen2.csv}{Model},
                table/col sep=semicolon,
                y dir=reverse,
                xticklabel={\pgfmathprintnumber{\tick}\%},
                xticklabel style={font=\tiny},
                yticklabel style={font=\tiny},
                axis line style={draw=none},
                tick style={draw=none},
                xmajorgrids=true,
                grid style={dotted, gray!50},
                legend style={
                    at={(0.5,-0.25)}, 
                    anchor=north,
                    legend columns=5,
                    draw=none,
                    font=\scriptsize
                },
            ]
                \addplot[fill=col5, draw=none] table [x=S5, y expr=\coordindex] {plots/Scen2.csv};
                \addplot[fill=col4, draw=none] table [x=S4, y expr=\coordindex] {plots/Scen2.csv};
                \addplot[fill=col3, draw=none] table [x=S3, y expr=\coordindex] {plots/Scen2.csv};
                \addplot[fill=col2, draw=none] table [x=S2, y expr=\coordindex] {plots/Scen2.csv};
                \addplot[fill=col1, draw=none] table [x=S1, y expr=\coordindex] {plots/Scen2.csv};
                \legend{Score 5, Score 4, Score 3, Score 2, Score 1}
            \end{axis}
        \end{tikzpicture}\vspace{-0.7em} 
        \caption{Scenario 2: Complex Reasoning}\label{fig:image2}
    \end{subfigure}\vspace{0.2em} 

    \begin{subfigure}{\columnwidth}
        \centering
        \begin{tikzpicture}
            \begin{axis}[
                xbar stacked,
                width=\textwidth, height=2.8cm,
                bar width=8pt,
                xmin=0, xmax=100,
                ytick=data,
                yticklabels from table={plots/Scen3.csv}{Model},
                table/col sep=semicolon,
                y dir=reverse,
                xticklabel={\pgfmathprintnumber{\tick}\%},
                xticklabel style={font=\tiny},
                yticklabel style={font=\tiny},
                axis line style={draw=none},
                tick style={draw=none},
                xmajorgrids=true,
                grid style={dotted, gray!50},
                legend style={
                    at={(0.5,-0.25)}, 
                    anchor=north,
                    legend columns=5,
                    draw=none,
                    font=\scriptsize
                },
            ]
                \addplot[fill=col5, draw=none] table [x=S5, y expr=\coordindex] {plots/Scen3.csv};
                \addplot[fill=col4, draw=none] table [x=S4, y expr=\coordindex] {plots/Scen3.csv};
                \addplot[fill=col3, draw=none] table [x=S3, y expr=\coordindex] {plots/Scen3.csv};;
                \legend{$\Delta$4, $\Delta$3, $\Delta$2}
            \end{axis}
        \end{tikzpicture}\vspace{-0.7em}
        \caption{Scenario 3: Deterministic Deletion (Delta Scores)}\label{fig:image3}
    \end{subfigure}\vspace{-0.2em}
    \caption{Judicial scores for \textbf{Scenario 1} and \textbf{2}, and $\Delta$ scores for \textbf{3}.}\vspace{-1em}
\end{figure}

\section{Conclusion}
We introduce \textsc{Ruva}, redefining Personal AI from an opaque service into a transparent framework. By adopting a neuro-symbolic GraphRAG paradigm, we addressed the "Black Box" limitations that plague current retrieval systems.
\textsc{Ruva} is viable on edge devices, achieving interactive latencies and high semantic accuracy. This proves that the cloud is no longer a prerequisite for intelligence, and data sovereignty is possible. Thanks to the graph-based topology, \textsc{Ruva} enables deterministic deletion and  operationalizes the "Right to be Forgotten".
\textsc{Ruva} returns agency to the human, enabling the user to inspect, correct, and erase the memories held by their AI. \textsc{Ruva} demonstrates that the most powerful AI is the one that the user can trust the most.

\bibliographystyle{named}
\bibliography{demo}



\end{document}